\documentclass{article}

\usepackage{arxiv}

\usepackage[utf8]{inputenc} 
\usepackage[T1]{fontenc}    
\usepackage{hyperref}       
\usepackage{url}            
\usepackage{booktabs}       
\usepackage{amsfonts}       
\usepackage{nicefrac}       
\usepackage{microtype}      
\usepackage[comma,authoryear]{natbib}
\usepackage{amsmath}
\usepackage{times}
\usepackage{latexsym}
\usepackage{graphicx}
\usepackage{algorithm}
\usepackage{algorithmicx}
\usepackage{algpseudocode}
\usepackage{amsmath}
\usepackage{multirow}
\usepackage{multicol}
\usepackage{setspace}
\usepackage{makecell}
\usepackage{amssymb}
\usepackage{diagbox}
\usepackage[shortlabels]{enumitem} \setlist[enumerate, 1]{1\textsuperscript{o}}

\title{Retrieving Event-related Human Brain Dynamics from Natural Sentence Reading}

\author{
  Xinping Liu\\
  School of ICT\\
  University of Tasmania\\
  Sandy Bay, TAS 7005 \\
  \texttt{xinpingl@utas.edu.au} \\
   \And
  Zehong Cao\\
  School of ICT\\
  University of Tasmania\\
  Sandy Bay, TAS 7005 \\
  \texttt{zehong.cao@utas.edu.au} \\
}

\begin{document}
\maketitle
\begin{abstract}
Electroencephalography (EEG) signals recordings when people reading natural languages are commonly used as a cognitive method to interpret human language understanding in neuroscience and psycholinguistics. Previous studies have demonstrated that the human fixation and activation in word reading associated with some brain regions, but it is not clear when and how to measure the brain dynamics across time and frequency domains. In this study, we propose the first analysis of event-related brain potentials (ERPs), and event-related spectral perturbations (ERSPs) on benchmark datasets which consist of sentence-level simultaneous EEG and related eye-tracking recorded from human natural reading experiment tasks. Our results showed peaks evoked at around 162 ms after the stimulus (starting to read each sentence) in the occipital area, indicating the brain retriving lexical and semantic visual information processing approaching 200 ms from the sentence onset. Furthermore, the occipital ERP around 200ms presents negative power and positive power in short and long reaction times. In addition, the occipital ERSP around 200ms demonstrated increased high gamma and decreased low beta and low gamma power, relative to the baseline. Our results implied that most of the semantic-perception responses occurred around the 200ms in alpha, beta and gamma bands of EEG signals. Our findings also provide potential impacts on promoting cognitive natural language processing models evaluation from EEG dynamics.
\end{abstract}

\section{Introduction} \label{intro}
Preserving human languages by text has been influencing the human cognitive growth, and it is also considered as the primary medium for preservation of natural language processing (NLP). However, research on how human brains retrieve and process the meaning from text-based languages is relatively few. Previous research \cite{Mason06} has used custom reading materials to explore particular facets of linguistic material such as event-related potentials (ERPs) and eye motions. Later on, using Electroencephalography (EEG) in conjunction with eye-tracking has become significant measures to study the temporal dynamics of natural sentence reading, because EEG signals have excellent time resolution and comparatively low costs to be recorded \cite{Dimigen11,Loberg18,HollensteinNZhange19}. For example, a recent study \cite{Pfeiffer20} indicated that text-based emotion analyze could be significantly enhanced by computing fixation-related potentials (FRPs) as one of ERPs on a neurocognitive dataset named ZuCo 1.0 \cite{Hollenstein18} that recorded from human readers during naturalistic reading. To measure event-related synchronisations, \cite{Roach2008} quantified the timeline of event-related spectral perturbations (ERSPs) that could generate brain dynamic changes in amplitude of the broadband EEG frequency spectrum.

However, the studies above only focused on wording or naive reading tasks without comprehensive event-related brain dynamic analyze, so it is still unknown how ERP and ERPS changes in human natural sentence reading. \textbf{In this work, we investigated EEG brain dynamics of ERPs and ERSPs through both channel and component forms for the total of 18 subjects during a recent natural sentence reading task, ZuCo 2.0} \citep{HollensteinZ219}. This task supports to analyse the differences in cognitive processing between natural reading, and also allows us to precisely extract the EEG signals for sentence-level processing and detecting human retrieving lexical and semantic information in natural sentence reading.

\begin{figure*}[htbp]
    \begin{minipage}{\linewidth}
    \includegraphics[width=\linewidth]{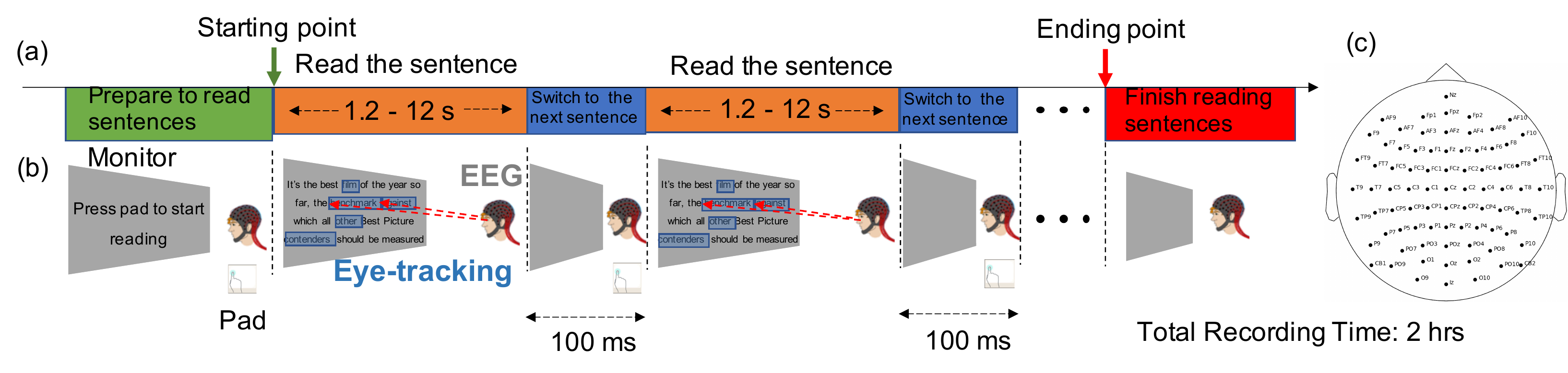}
    \caption{Natural Sentence Reading Experiment. (a) Natural reading phases. (b) Each sentence onset and finished state and switching to the next sentence by control pad. The sentence reading and sentence switching were performed alternately, and each sentence switching lasted 100 ms, while sentence reading lasted from 1.2 second to 12 seconds. (c) Location map of multi-channel EEG. }
    \label{figpdm}
    \end{minipage}
\end{figure*}

\section{Related Work}

\textbf{ERPs} \cite{Frank13,Frank15} are electrical potential responses or reaction measured in relation to an event, such as reading each sentence in our study. For example, N200 is a negative wave in connection with a baseline because it occurs at around 200 to 350 ms after the onset of the first word in a sentence, but it may cause some mismatch issues after an infrequent change in the case of visual stimuli \cite{Brattico2006}. The recent study \cite{Pfeiffer20} has indicated that text-based emotion analyze can be significantly enhanced by using FRPs to analyze human readers during naturalistic reading, but it only focused on the potentials of word-based reading. Also, \cite{Hale18, Sharmistha19} applied ERPs as a neuroimage feature to retrieve the lexical and semantic information from the natural language content for designing cognitive word embedding evaluation in NLP, but still, there has some significant loss possibly due to not precise patterns. In terms of \textbf{ERSPs}, one study \cite{zhang2017investigation} investigated emotion and working memory analyze to reading, but it did not touch the natural sentence reading tasks.

\section{Method}
\subsection{Experiment Paradigm and Data}

The natural sentence reading experiment and data was performed and collected in ZuCo 2.0 \citep{HollensteinZ219}, which recorded from 18 healthy subjects who are all native  English speakers originating from Canada, USA, UK or Australia. The sentences were displayed one by one at the same position on the monitor during the natural reading paradigm. The text was displayed on a light grey backdrop in black with font size 20-point Arial, resulting in a letter height of 0.8 mm or 0.674 $^{\circ}$. During the entire reading task, a sum of 80 letters or 13 words were presented per line, and the subjects who participate in this natural reading experiment read each sentence at their own speed. 

In the task, the 349 sentences were displayed one by one at the same position on the monitor during the natural reading task for each subject. The recording devices include both 128-channel EEG and eye-tracking that used to collect brain signals and eye motions when subjects conducted natural sentence reading, as shown in \textbf{Fig. 1}. In this work, one trial was considered as reading each sentence, and there are a total of 6,282 trials (18 subjects multiplied by 349 sentences in the natural reading paradigm). 

\subsection{Data Processing and analyze}
\paragraph{EEG Processing} The raw EEG data were loaded by EEGLAB \citep{EEGLAB04} with the removal of bad electrodes, keeping 105 EEG channels. We then removed electromyogram (EMG) and electrooculogram (EOG) artifacts and minimised other artifacts by setting a finite impulse response filter 1-50Hz. Next, we conducted the EEG electrodes re-referencing by the vertex electrode Cz, and the re-referenced EEG data were measured by Z Score formula normalisation. Furthermore, we conducted the data segmentation to extract all trials where two time-locking events (sentence onset event and sentence finished event) were selected. What is more, the trial limits were 100 ms before starting sentence reading (-100 ms, act as the baseline) to 1200 ms after the reading event (+1200 ms), because the sentence interval sets 100 ms and the minimum interval between sentence onset and sentence finished is 1223 ms.

In addition, to represent synchronous brain activities, decomposing the pre-processed data by Independent Component analyze (ICA) is an effective approach to investigate the diversity of source information typically contained in EEG data. In this study, we merged a total of 18 pre-processed single-subject EEG data and conducted the statistical comparisons by running the merged-subjects ICA decomposition to find the potential similar independent component sources across 18 subjects.

\paragraph{EEG analyze: ERP/ERSP in channel-based and component-based forms}

The 1-D event latency scalp maps and 2-D ERP/ERSP images could produce to intuitively discover the brain dynamics of human natural sentence reading in both forms of channel-based and component-based EEG patterns, respectively. In terms of ERPs, to analyse the channel-based ERPs, we chose the most relevant channel to the corresponding events from the scalp map where the power is concentrated among a total of 18 subjects. Meanwhile, to analyse component-based ERPs, we calculated and selected the power spectrum for isolated components to identify which component contributes the most to the ERPs. For ERSPs, similarly, we determined the channel which is most relevant to the corresponding events and analysed spectral perturbations, and then calculated and selected a component that the source contributes the most to the task. Please note that the significance level sets 0.01 with removing the baseline for plotting the ERSPs in this study.

\section{Results}

Our results show the comprehensive group analyze and visualisation of 18 subjects of EEG dynamics for natural sentence reading, including 1-D event latency potentials plots and 2-D ERP and ERSP images in channel-based and component-based forms.

\paragraph{Channel/Component 1-D event latency potentials} 

\begin{figure}[!htbp]
    \begin{minipage}{\columnwidth}
    \includegraphics[width=\linewidth]{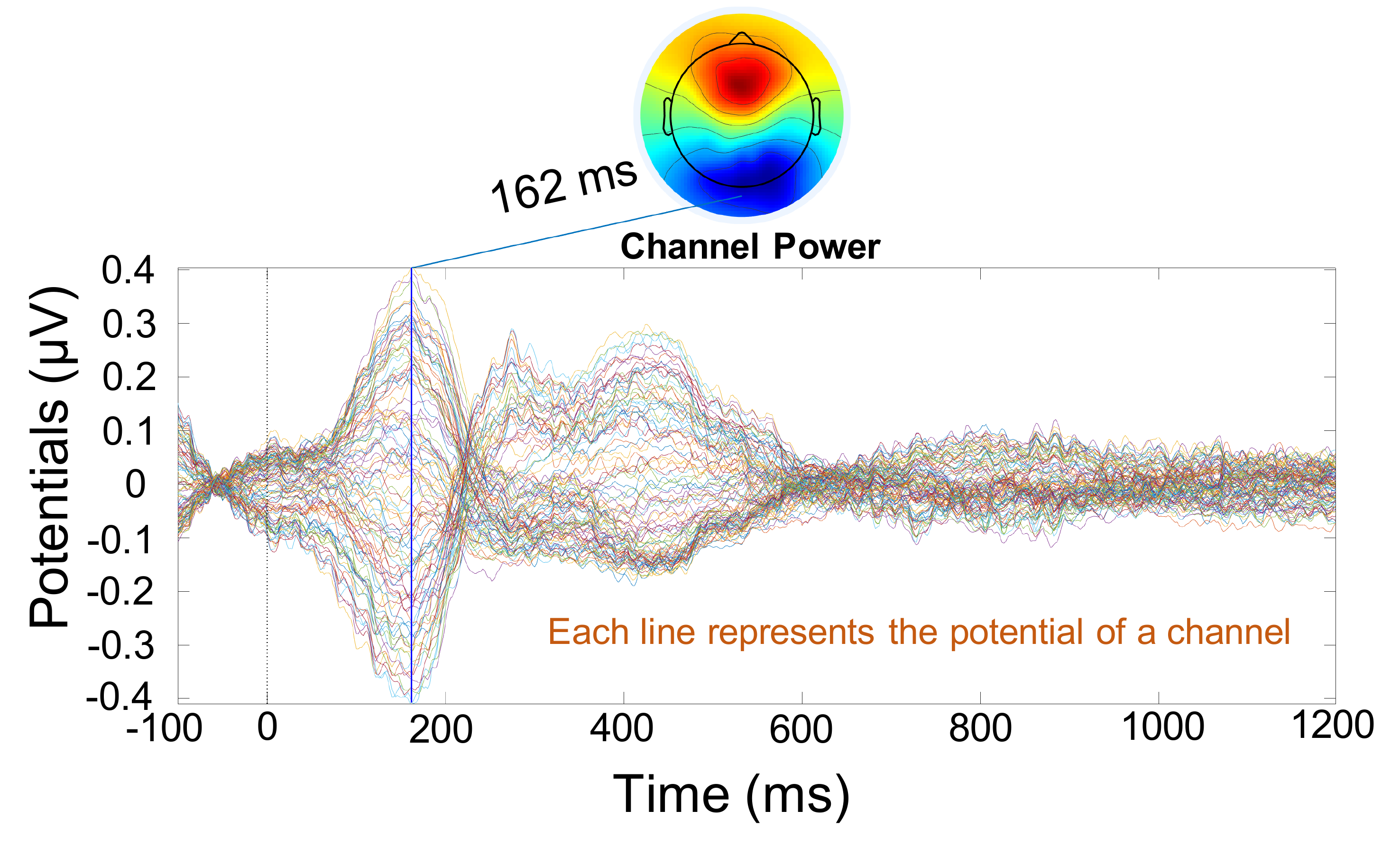}
    \caption{Channel-based 1-D event latency potientials with the scalp map among a total of 18 subjects for natural sentence reading. }
    \label{figlan}
    \end{minipage}
\end{figure}

As shown in \textbf{Fig.~\ref{figlan}}, from 1-D event latency potentials among all subjects, we found highest peaks at 162 ms from the onset natural sentence reading, which may indicate that a significant EEG dynamic (particular in reduced power in the occipital region) approaching 200 ms (N200) and this finding also supported by evidence for the early latency of lexical and semantic information retrieval in visual word recognition \citep{Hauk12,Pfeiffer20}. Thus, we assume that the human start retrieves lexical and semantic information near-simultaneously at the central occipital region (Oz channel) within 200 ms of the sentence reading onset. 

\begin{figure}[!htbp]
    \begin{minipage}{\columnwidth}
    \includegraphics[width=\linewidth]{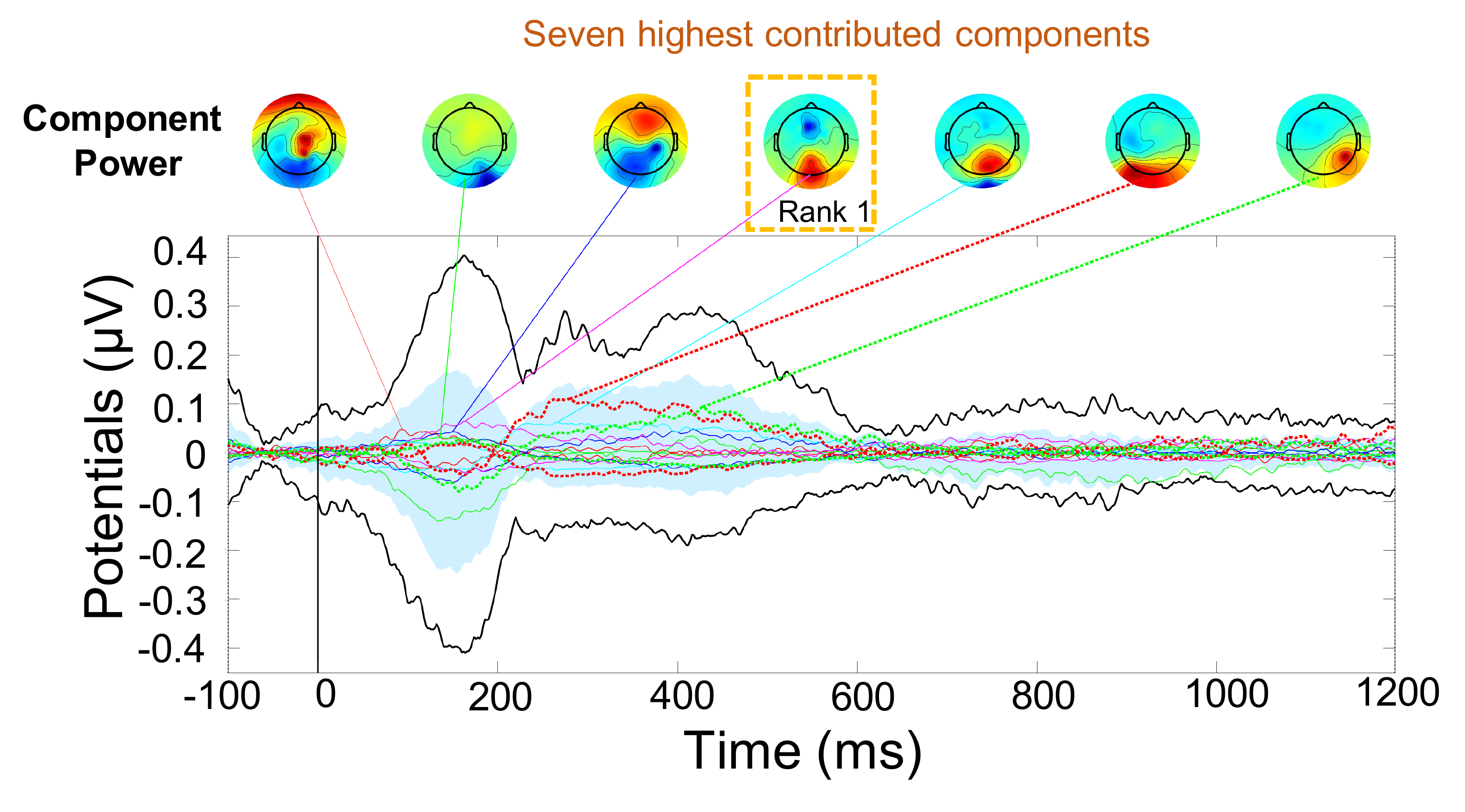}
    \caption{Components-based 1-D event latency potientials with the scalp map among a total of 18 subjects for natural sentence reading.}
    \label{figcomp}
    \end{minipage}
\end{figure}

Following above, we also extracted 7 most significant component maps from 1-D event latency potentials, as shown in \textbf{Fig.~\ref{figcomp}}, which showed the occipital component contributed most to the task (marked as rank 1). The results in the section are used for determining that the occipital region is the most contribution area related to the natural sentence reading task, because the occipital results displayed the strongest response in the text-based retrieval of sentiment information than that of the other regions of the brain.

\paragraph{Channel/Component 2-D ERP Results} 

\begin{figure}[!htbp]
    \begin{minipage}{\linewidth}
    \includegraphics[width=\linewidth]{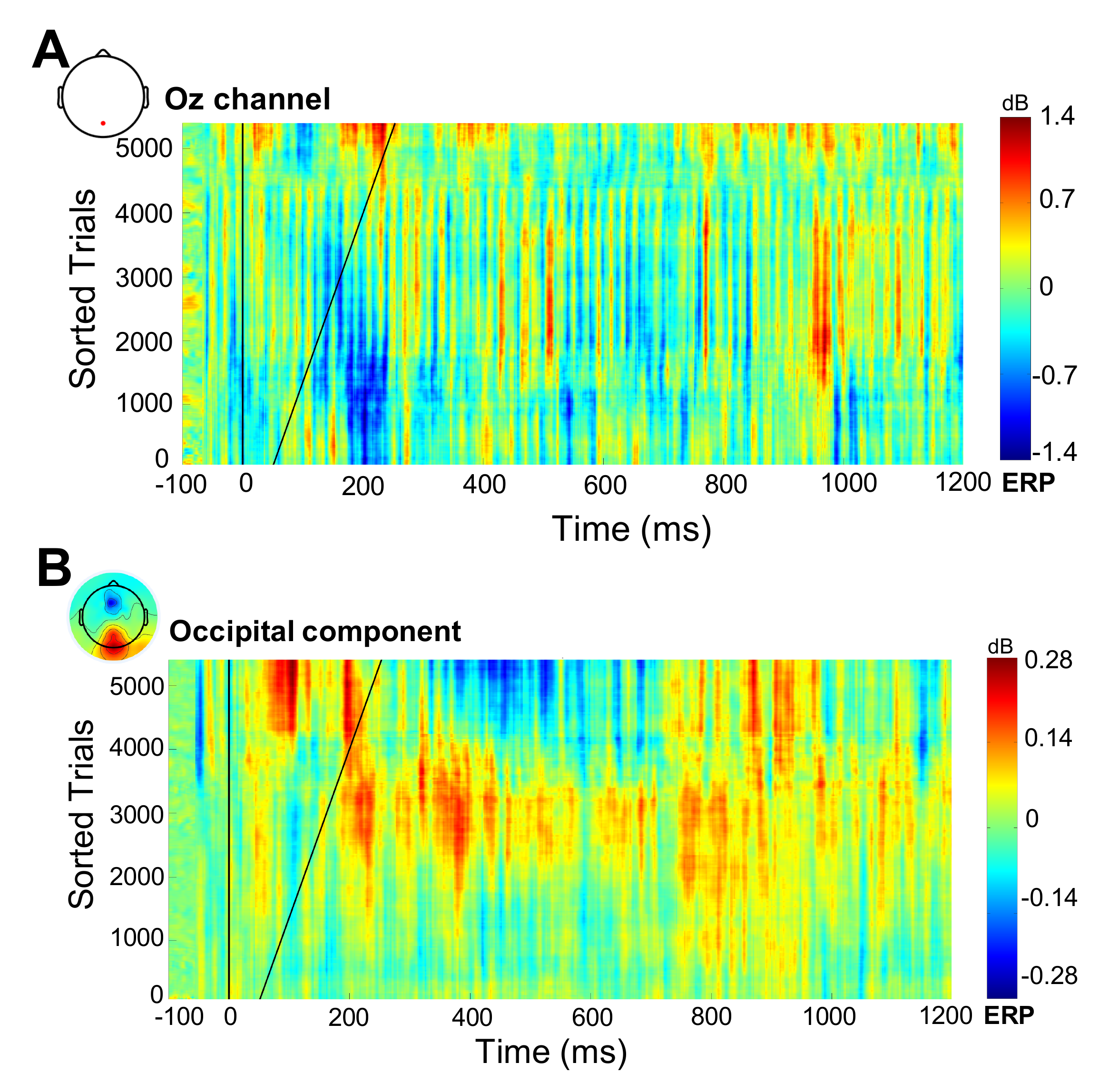}
    \caption{2-D ERP results for natural sentence reading in channel-based (Oz) and component-based (occipital) forms. }
    \label{figerp}
    \end{minipage}
\end{figure}

As shown in \textbf{Fig.~\ref{figerp}-A}, we located the channel Oz as the central occipital region to plot channel-based ERP results, which present the negative power (marked as the blue colour) around 200 ms in short reaction time (between 1-2000 trials). However, from the occipital component-based ERP image as shown in \textbf{Fig.~\ref{figerp}-B}, we observed the positive power (marked as the red colour) around 200 ms in medium (between 2000-4000 trials) and long (between 4000-6000 trials) reaction times. This is a new finding that may be beneficial for retrieving the diverse patterns of brain visual reading processing.

\paragraph{Channel/Component 2-D ERSP Results} 

\begin{figure}[!htbp]
    \begin{minipage}{\linewidth}
    \includegraphics[width=\linewidth]{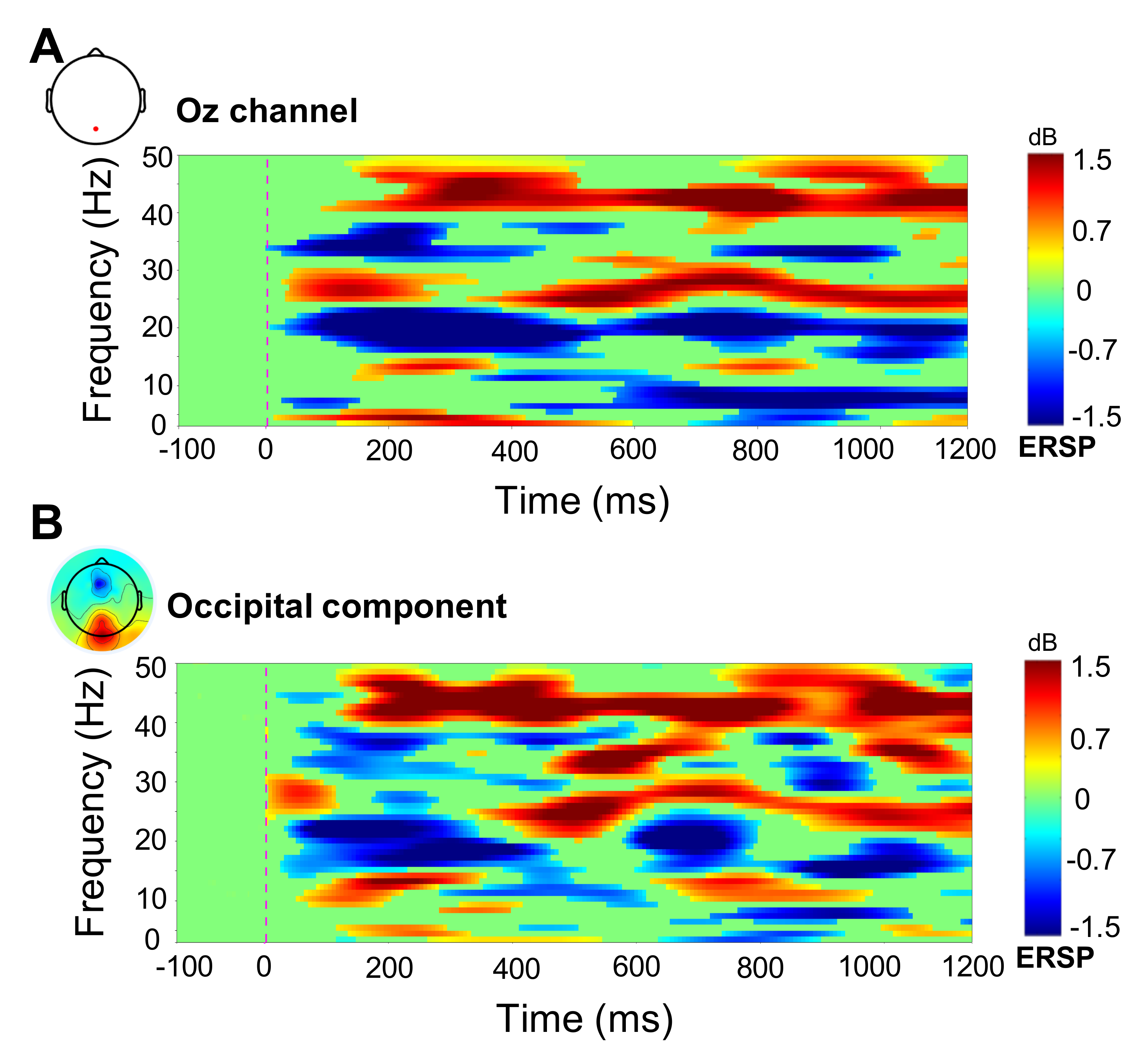}
    \caption{ 2-D ERSP results for natural sentence reading in channel-based (Oz) and component-based (occipital) forms.}
    \label{figersp}
    \end{minipage}
\end{figure}

In terms of the channel-based and component-based ERSPs in the occipital region, they demonstrated a similar performance around 200 ms, as shown in \textbf{Fig.~\ref{figersp}}. The Oz-channel ERSP as presented in \textbf{Fig.~\ref{figersp}-A}, at the time of around 200ms, demonstrated increased high alpha (10-12Hz), high beta (25-30Hz), and high gamma (40-50Hz) power, and decreased low beta (13-25Hz) and low gamma (30-40Hz) power, relative to the baseline (the interval of sentence reading). In terms of the occipital component-channel ERSP around 200 ms, \textbf{Fig.~\ref{figersp}-B} presented increased high alpha (10-12Hz) and high gamma (40-50Hz) power, and decreased low beta (13-25Hz) and low gamma (30-40Hz) power, relative to the baseline (the interval of sentence reading).

Furthermore, ERSP results in the occipital region around 200 ms primarily consist with the longer latencies covering from 300 ms to 1200 ms (the ending point of each sentence reading), particularly in the enhanced high gamma (40-50Hz) and the reduced low beta (13-25Hz) and low gamma (30-40Hz) power spectral, relative to the baseline, suggesting the patterns for retrieving lexical and semantic information in the natural sentence reading task.

\section{Conclusion}
Our study successfully identified the spatiotemporal neural dynamics of sentiment processing during the naturalistic reading of English sentences. Combining high-density EEG and eye-tracking data, we provide clear evidence that essential processes of lexical and semantic information retrieval occur around the 200 ms of each sentence onset, by retrieving EEG patterns of negative power and positive power in short and long reaction times and increased high gamma and decreased low beta and low gamma power. We believe that these EEG dynamics found from this study for human language understanding could benefit for semantic representations of sentence reading and cognitive evaluation modelling for NLP.

\clearpage

\bibliographystyle{unsrtnat}  
\bibliography{references}

\begin{thebibliography}{15}
\providecommand{\natexlab}[1]{#1}
\providecommand{\url}[1]{\texttt{#1}}
\expandafter\ifx\csname urlstyle\endcsname\relax
  \providecommand{\doi}[1]{doi: #1}\else
  \providecommand{\doi}{doi: \begingroup \urlstyle{rm}\Url}\fi

\bibitem[Mason and Just(2006)]{Mason06}
Robert~A. Mason and Marcel~Adam Just.
\newblock Chapter 19 - neuroimaging contributions to the understanding of
  discourse processes.
\newblock In Matthew~J. Traxler and Morton~A. Gernsbacher, editors,
  \emph{Handbook of Psycholinguistics (Second Edition)}, pages 765 -- 799.
  Academic Press, London, second edition edition, 2006.
\newblock ISBN 978-0-12-369374-7.
\newblock \doi{https://doi.org/10.1016/B978-012369374-7/50020-1}.
\newblock URL
  \url{http://www.sciencedirect.com/science/article/pii/B9780123693747500201}.

\bibitem[Dimigen et~al.(2011)Dimigen, Sommer, Hohlfeld, Jacobs, and
  Kliegl]{Dimigen11}
Olaf Dimigen, Werner Sommer, Annette Hohlfeld, Arthur~M Jacobs, and Reinhold
  Kliegl.
\newblock Coregistration of eye movements and eeg in natural reading: analyses
  and review.
\newblock \emph{Journal of experimental psychology. General}, 140\penalty0
  (4):\penalty0 552—572, November 2011.
\newblock ISSN 0096-3445.
\newblock \doi{10.1037/a0023885}.
\newblock URL \url{https://doi.org/10.1037/a0023885}.

\bibitem[Loberg et~al.(2018)Loberg, Hautala, Hämäläinen, and
  Leppänen]{Loberg18}
Otto Loberg, Jarkko Hautala, Jarmo~A. Hämäläinen, and Paavo H.~T. Leppänen.
\newblock Semantic anomaly detection in school-aged children during natural
  sentence reading – a study of fixation-related brain potentials.
\newblock \emph{PLoS ONE}, 13\penalty0 (12):\penalty0 e0209741, 2018.
\newblock \doi{10.1371/journal.pone.0209741}.
\newblock URL \url{https://app.dimensions.ai/details/publication/pub.1110936695
  and
  https://journals.plos.org/plosone/article/file?id=10.1371/journal.pone.0209741&type=printable}.

\bibitem[Hollenstein and Zhang(2019)]{HollensteinNZhange19}
Nora Hollenstein and Ce~Zhang.
\newblock Entity recognition at first sight improving {NER} with eye movement
  information.
\newblock \emph{CoRR}, abs/1902.10068, 2019.
\newblock URL \url{http//arxiv.org/abs/1902.10068}.

\bibitem[Pfeiffer et~al.(2020)Pfeiffer, Hollenstein, Zhang, and
  Langer]{Pfeiffer20}
Christian Pfeiffer, Nora Hollenstein, Ce~Zhang, and Nicolas Langer.
\newblock Neural dynamics of sentiment processing during naturalistic sentence
  reading.
\newblock \emph{NeuroImage}, 218:\penalty0 116934, 2020.
\newblock ISSN 1053-8119.
\newblock \doi{https://doi.org/10.1016/j.neuroimage.2020.116934}.
\newblock URL
  \url{http://www.sciencedirect.com/science/article/pii/S1053811920304201}.

\bibitem[Hollenstein et~al.(2018)Hollenstein, Rotsztejn, Troendle, Pedroni,
  Zhang, and Langer]{Hollenstein18}
Nora Hollenstein, Jonathan Rotsztejn, Marius Troendle, Andreas Pedroni,
  Ce~Zhang, and Nicolas Langer.
\newblock Zuco, a simultaneous eeg and eye-tracking resource for natural
  sentence reading.
\newblock \emph{Scientific Data}, 5, 2018.

\bibitem[Roach and Mathalon(2008)]{Roach2008}
Brian~J. Roach and Daniel~H. Mathalon.
\newblock Event-related eeg time-frequency analysis: an overview of measures
  and an analysis of early gamma band phase locking in schizophrenia.
\newblock \emph{Schizophrenia bulletin}, 34\penalty0 (5):\penalty0 907--926,
  Sep 2008.

\bibitem[Hollenstein et~al.(2020)Hollenstein, Troendle, Zhang, and
  Langer]{HollensteinZ219}
Nora Hollenstein, Marius Troendle, Ce~Zhang, and Nicolas Langer.
\newblock {Z}u{C}o 2.0: A dataset of physiological recordings during natural
  reading and annotation.
\newblock In \emph{Proceedings of the 12th Language Resources and Evaluation
  Conference}, pages 138--146, Marseille, France, May 2020. European Language
  Resources Association.
\newblock ISBN 979-10-95546-34-4.
\newblock URL \url{https://www.aclweb.org/anthology/2020.lrec-1.18}.

\bibitem[Frank et~al.(2013)Frank, Fernandez~Monsalve, Thompson, and
  Vigliocco]{Frank13}
Stefan~L. Frank, Irene Fernandez~Monsalve, Robin~L. Thompson, and Gabriella
  Vigliocco.
\newblock Reading time data for evaluating broad-coverage models of english
  sentence processing.
\newblock \emph{Behavior Research Methods}, 45\penalty0 (4):\penalty0
  1182--1190, 2013.

\bibitem[Frank et~al.(2015)Frank, Otten, Galli, and Vigliocco]{Frank15}
Stefan~L. Frank, Leun~J. Otten, Giulia Galli, and Gabriella Vigliocco.
\newblock The erp response to the amount of information conveyed by words in
  sentences.
\newblock \emph{Brain and Language}, 140:\penalty0 1--11, 2015.

\bibitem[Brattico et~al.(2006)Brattico, Tervaniemi, Naatanen, and
  Peretz]{Brattico2006}
Elvira Brattico, Mari Tervaniemi, Risto Naatanen, and Isabelle Peretz.
\newblock Musical scale properties are automatically processed in the human
  auditory cortex.
\newblock \emph{Gene Expression Patterns}, 1117:\penalty0 162--174, October
  2006.
\newblock ISSN 1567-133X.
\newblock \doi{10.1016/j.brainres.2006.08.023}.

\bibitem[Hale et~al.(2018)Hale, Dyer, Kuncoro, and Brennan]{Hale18}
John Hale, Chris Dyer, Adhiguna Kuncoro, and Jonathan Brennan.
\newblock Finding syntax in human encephalography with beam search.
\newblock In \emph{Proceedings of the 56th Annual Meeting of the Association
  for Computational Linguistics (Volume 1: Long Papers)}, pages 2727--2736,
  Melbourne, Australia, July 2018. Association for Computational Linguistics.
\newblock \doi{10.18653/v1/P18-1254}.
\newblock URL \url{https://www.aclweb.org/anthology/P18-1254}.

\bibitem[Jat et~al.(2019)Jat, Tang, Talukdar, and Mitchell]{Sharmistha19}
Sharmistha Jat, Hao Tang, Partha~P. Talukdar, and Tom~Michael Mitchell.
\newblock Relating simple sentence representations in deep neural networks and
  the brain.
\newblock \emph{CoRR}, abs/1906.11861, 2019.
\newblock URL \url{http//arxiv.org/abs/1906.11861}.

\bibitem[Delorme and Makeig(2004)]{EEGLAB04}
Arnaud Delorme and Scott Makeig.
\newblock Eeglab: an open source toolbox for analysis of single-trial eeg
  dynamics including independent component analysis.
\newblock \emph{Journal of neuroscience methods}, 134\penalty0 (1):\penalty0
  9—21, March 2004.
\newblock ISSN 0165-0270.
\newblock \doi{10.1016/j.jneumeth.2003.10.009}.
\newblock URL \url{https://doi.org/10.1016/j.jneumeth.2003.10.009}.

\bibitem[Hauk et~al.(2012)Hauk, Coutout, Holden, and Chen]{Hauk12}
O.~Hauk, C.~Coutout, A.~Holden, and Y.~Chen.
\newblock The time-course of single-word reading: Evidence from fast behavioral
  and brain responses.
\newblock \emph{NeuroImage}, 60\penalty0 (2):\penalty0 1462--1477, 2012.
\newblock \doi{10.1016/j.neuroimage.2012.01.061}.
\newblock URL \url{https://doi.org/10.1016/j.neuroimage.2012.01.061}.

\end{thebibliography}


\end{document}